\documentclass{article}

\usepackage{arxiv}

\usepackage[T1]{fontenc}
\usepackage[utf8]{inputenc}
\usepackage{times}
\usepackage{microtype}
\usepackage{inconsolata}
\usepackage{graphicx}
\usepackage{amsmath}
\usepackage{tikz}
\usetikzlibrary{arrows.meta, positioning, shapes.geometric, backgrounds, fit, calc}

\usepackage[authoryear,round]{natbib}
\let\cite\citep
\usepackage{url}
\usepackage[hidelinks,breaklinks]{hyperref}
\usepackage{orcidlink}
\title{Self-Consistency of Entity Sets and Citations in a Deep Research System under Repeated Prompting}
\renewcommand{\shorttitle}{Self-Consistency in a Deep Research System}
\date{}

\author{
  \normalfont\Large Lauri Pykälä\,\textsuperscript{\orcidlink{0009-0000-7807-4155}}
  and Kimmo K. Kaski\,\textsuperscript{\orcidlink{0000-0002-3805-9687}} \\[1em]
  \normalfont\large Department of Computer Science \\
  \normalfont\large Aalto University School of Science, P.O. Box 11000, 00076 Aalto, Finland \\
  \normalfont\large Correspondence: \href{mailto:lauri.pykala@aalto.fi}{\texttt{lauri.pykala@aalto.fi}}
}

\hypersetup{
  pdftitle={Self-Consistency of Entity Sets and Citations in a Deep Research System under Repeated Prompting},
  pdfauthor={Lauri Pykälä; Kimmo K. Kaski}
}

\begin{document}

\maketitle

\begin{abstract}
	Deep research agents (DRA) are commonly used for open-ended synthesis tasks where closed-world QA benchmarks are poorly suited to characterize relevant aspects of performance.
	We study the \emph{self-consistency} of one commercial deep research system under repeated identical prompting, representing each report with the set of named entities it mentions, categorised by a custom schema, and the set of citations it retrieves.
	Across repeated runs of a single open-ended cybersecurity related prompt, we quantify pairwise overlap to characterize how stable the retrieved and synthesized output structure is under repeated identical prompts.
    In ten runs, the pairwise entity Jaccard results in a median of $0.12$ and the pairwise citation Jaccard is below $0.02$, with discovery curves that do not visibly saturate.
	The magnitude and structure of (dis)agreement between entity sets varies by entity class, which suggests that self-consistency in a DRA is better understood as a function of the underlying entity space than as a stable property of the system.
\end{abstract}

\section{Introduction and Related Work}

Retrieval-Augmented Generation (RAG) is an umbrella term for methods designed to inject external knowledge into the context of a large language model at runtime, with the goal of providing more accurate, grounded, or updated information to improve the reliability and transparency of the answers it generates.
As suggested by the acronym, it has three main components: \emph{Retrieval} of data from external sources, \emph{Augmenting} the current query context with the most relevant information, and \emph{Generating} an answer that combines the Large Language Model's (LLM's) pre-trained knowledge with the retrieved information.

\citet{singhAgenticRetrievalAugmentedGeneration2025} suggest a taxonomy of RAG ranging from a simple keyword-based Na\"ive RAG through to more advanced Agentic RAG systems, known as deep research agents (DRAs) that can autonomously examine hundreds of online sources per query and produce long-form reports with citations in response to arbitrary requests.
Recent surveys characterise four core stages in the deep research pipeline, i.e. planning, question development, web exploration, and report generation, each of which admits substantial design variation across commercial implementations~\cite{zhangDeepResearchSurvey2025}.

Existing evaluations of deep research agents often emphasize claim-level factual accuracy, leaving aspects such as their handling of multi-source synthesis and the structural properties of the knowledge these systems produce less examined~\cite{huangDeepResearchAgents2025}.
Because the outputs are long-form and open-ended, being based on a web-scale retrieval step rather than a fixed corpus, their real-world usage is poorly characterized by the closed-world QA style benchmarks.
More recent evaluations have developed approaches such as decomposing reports into atomic facts~\cite{minFActScoreFinegrainedAtomic2023} and using metrics such as recall and relevance in addition to binary correctness~\cite{guptaDeepSearchQABridgingComprehensiveness2026}.

Large language models are inherently stochastic, and due to their autoregressive nature, the effects of small perturbations in their outputs compound over the sequences generated. \citet{zhaiEvaluatingStochasticityDeep2026} model the stochasticity of answers, findings, and citations in DRA using Markov information acquisition decision processes. Most prior treatments of DRA stochasticity focus on variability in answers or final claims, but the structural question of how stable the set of retrieved and synthesized referents is across repeated runs has received comparatively little attention.
Another line of work evaluates long-form generation by decomposing outputs into discrete units (atomic facts, claims, or named entities) and reasoning about coverage and overlap at that level rather than at the level of full text \cite{minFActScoreFinegrainedAtomic2023}.

Next, the stability of DRA under repeated prompting and decomposition of its output into discrete units is presented as the following research questions:

\begin{itemize}
	\item How consistent is a deep research system's entity coverage across repeated runs of the same prompt?
	\item How consistent is its set of cited sources across repeated runs?
	\item How consistent is the joint structure of \emph{which entities are grounded in which sources} across repeated runs?
\end{itemize}
Our approach contributes to addressing a gap in long-form LLM evaluation by examining how stable a single system's retrieved and synthesized outputs, operationalized through entity extraction, are under repeated identical prompting, rather than whether individual statements are verifiably factually correct.
The contributions presented here are threefold. First, we propose a three-level representation of DRA outputs, i.e.
canonicalized named entities, citations, and entity--citation pairs,
as an evaluation lens for self-consistency under repeated identical prompting.
Second, applying this lens to a single commercial DRA on an open-ended topical cybersecurity related prompt, we show that entity-level agreement is substantially higher than citation-level agreement.
Third, we show that pairwise agreement varies sharply across entity classes in a way that tracks the presumed size of the underlying entity space.

\section{Method}

\begin{figure}[!t]
\centering
\begin{tikzpicture}[
    font=\small,
    node distance=6mm,
    stage/.style={
        rectangle, rounded corners=2pt,
        draw=black!70, fill=blue!5,
        minimum height=9mm, minimum width=20mm,
        align=center, inner sep=2pt
    },
    promptbox/.style={
        rectangle, rounded corners=2pt,
        draw=black!70, fill=yellow!10,
        minimum height=14mm, text width=28mm,
        align=left, inner sep=4pt,
        font=\scriptsize\ttfamily
    },
    runcard/.style={
        rectangle, rounded corners=1pt,
        draw=black!70, fill=blue!10,
        minimum height=14mm, minimum width=16mm,
        align=center, inner sep=1pt,
        font=\scriptsize
    },
    arr/.style={-{Latex[length=2mm]}, thick},
    tool/.style={font=\scriptsize\itshape, text=black!60}
]

\node[promptbox] (prompt) {``Write a report on the state of the art in LLM-aided vulnerability discovery and exploit development. What are the key companies, products and services involved? How have nation states and other actors reacted?''};

\node[runcard, right=10mm of prompt] (run3) {};
\node[runcard, fill=blue!10] at ([shift={(-1mm,1mm)}]run3) (run2) {};
\node[runcard, fill=blue!15] at ([shift={(-1mm,1mm)}]run2) (run1)
    {\textit{entities} \\[-0.5mm] \rule{12mm}{0.3pt} \\[-0.5mm] \textit{citations}};
\node[above=1mm of run1, font=\scriptsize\bfseries] {$\times 10$ runs};

\node[stage, right=12mm of run3] (extract) {Extract\\entities\\\& citations};

\node[stage, right=of extract] (canon) {Canonicalize\\(3$\times$ vote)};

\node[stage, right=of canon] (analyze) {Jaccard\\\& discovery\\curves};

\draw[arr] (prompt.east) -- (run1.west);
\draw[arr] (run3.east) -- (extract.west);
\draw[arr] (extract) -- (canon);
\draw[arr] (canon) -- (analyze);

\node[tool, below=1mm of run3.south] {Gemini Deep Research};
\node[tool, below=1mm of extract.south] {langextract};
\node[tool, below=1mm of canon.south] {gabriel};
\node[tool, below=1mm of analyze.south] {};

\end{tikzpicture}
\caption{\textbf{Analysis pipeline}. We query Google Deep Research ten times with the same cybersecurity-related prompt and treat each generated report as one run. For each run, we extract entities and citation URLs using langextract, canonicalize entity strings by class using a three-way voting procedure with gabriel, and compute pairwise Jaccard similarities and discovery curves over the resulting canonical entity and citation sets.}
\label{fig:pipeline}
\end{figure}
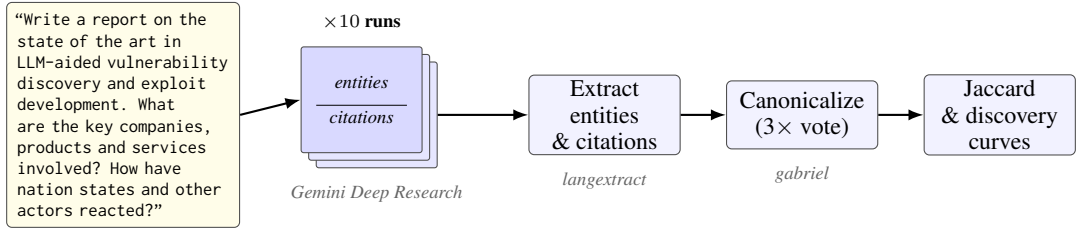

To study the self-consistency of a deep research system under fixed user inputs, we issue an identical open-ended cybersecurity prompt to a single commercial deep research system (Google \texttt{deep-research-preview-04-2026}, accessed through the API on 2026-04-27) in $10$ repeated iterations of the same prompt, shown in Figure~\ref{fig:pipeline}.
Each of the ten runs produced a main report with links to citations.
Inline citations in the raw response were emitted by the agent as a redirect URL. These URLs are followed by
an HTTP GET request with a Chrome-like \texttt{User-Agent} and the resolved URL is recorded.
These raw reports and cited sources are cleaned with a deterministic preprocessing pipeline before extraction. Resolved citation URLs were retained as citation identifiers, but the content of cited web pages was not included in the entity-overlap analysis.

Entities were extracted from all preprocessed documents using the \texttt{langextract} library \footnote{\texttt{langextract} is not an officially supported Google product.} with Google Gemini 2.5-flash as the extraction model \cite{langextract111}.
Raw entity surface forms were canonicalized using the \texttt{openai-gabriel} library, a general-purpose information extraction library that has been described and validated in \citet{asirvathamGPTMeasurementTool2026}.
Deduplication was run separately for each of the eleven entity classes.

The method operates in two stages: first, embedding-based grouping clusters similar surface forms into groups of up to 500 that are then assigned labels in-context.
The model used was \texttt{gpt-5.4-mini}.
The process was repeated three times (\texttt{n\_runs=3}) with voting to reduce the effect of random noise.
The system prompt used was the default one provided with the package.

To quantify the stochasticity of the DRA reports, we compute pairwise Jaccard similarity over all $\binom{n}{2} = \binom{10}{2} = 45$ unordered pairs of runs. For two runs $A$ and $B$ with item sets $S_A$ and $S_B$,
$$J(A, B) = \frac{|S_A \cap S_B|}{|S_A \cup S_B|},$$
with $J = 0$ when both sets are empty. The pairwise Jaccard similarity was calculated for the following:

\begin{itemize}
	\item \textbf{Overall entities.} $S = \{(c, e)\}$, where $c$ is an extraction class and $e$ is a deduplicated canonical entity mentioned in the report, across all classes jointly.
	\item \textbf{Per-class entities.} For each extraction class $c$, $S_c = \{ e \}$ restricted to that class. This is done to disentangle the degree of variance across different entity classes.
	\item \textbf{Citations.} $S = \{u\}$ over distinct URLs.
    \item \textbf{Entity–citation edges.} $S=\{(c,e,u) \} $, where $e$ is a canonical entity of class $c$, and $u$ is the resolved citation URL assigned to that entity mention by the nearest following inline citation marker in the generated report. Edge overlap is therefore a strict URL-level proxy for whether the same entity is grounded in the same cited source across runs.

\end{itemize}

Pairwise Jaccard describes the degree of \emph{agreement between runs} but obfuscates the \emph{directional trajectory} of the separate runs. To investigate the latter, discovery curves are used. Given a fixed reading order $\pi = (d_1, \dots, d_n)$ of the $n$ runs and a per-document item set $X_d$, the discovery curve is as follows:

$$C_\pi(k) = \left| \bigcup_{i=1}^{k} X_{d_i} \right|, \quad k = 1, \dots, n,$$
i.e. the cumulative count of distinct items (entities or citations) seen after reading the first $k$ documents.
To reduce dependence on an arbitrary run ordering, we compute this curve over 1000 random permutations of the ten runs and report the mean accumulation curve in Figure~\ref{fig:discovery}.
Flattening curves indicate that repeated runs are converging on a stable set; curves that remain steep indicate that each additional run keeps retrieving unseen entities.

\section{Results}

Across the 10 runs, outputs contained a median of 5,292 words (range: 4,837--6,152), 199 extracted entity mentions (162--271), 162 distinct canonical entities after deduplication (139--193), 2.96 entities per 100 words (2.42--3.64), and 79 distinct citation URLs (66--101).

\begin{figure}[t]
	\centering
	\includegraphics[width=0.70\linewidth]{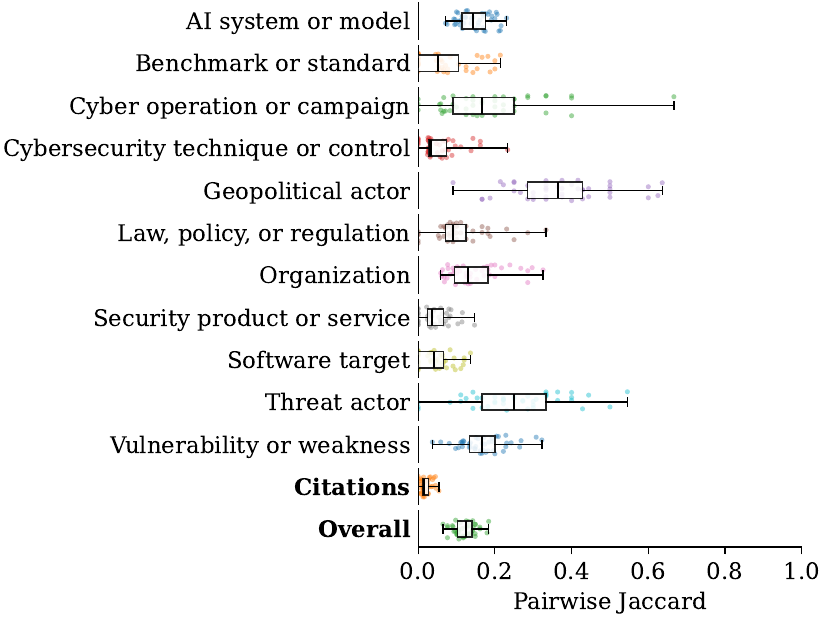}
	\caption{\textbf{Pairwise Jaccard similarity across ten repeated runs.} Each point represents one unordered pair of reports, for 45 run pairs in total. Rows show Jaccard similarity over canonical entities within each class, resolved citation URLs, and the overall canonical entity set. Boxplots summarize the distribution across run pairs. Overall entity overlap is low, with median Jaccard approximately 0.12, while citation overlap is much lower, with median citation Jaccard just over 0.01. Entity–citation edge overlap is omitted from the plot because its median Jaccard is similarly near zero and visually indistinguishable from the citation row.}
	\label{fig:pairwise-strip}
\end{figure}

Figure~\ref{fig:pairwise-strip} shows the distribution of pairwise Jaccard similarities across all 45 run pairs, broken down by entity class as well as aggregated for the overall entity set and for cited sources.
Agreement between any two runs is low in absolute terms: the overall entity Jaccard has a median of $0.12$, meaning that any two reports share only about one canonical entity in eight, despite being responses to an identical prompt.
Overlap of citations is even lower, with the median Jaccard similarity score between sets of citations in two reports being just over $0.01$, indicating that repeated runs of the same query rarely cite the same URLs.
We additionally computed pairwise Jaccard similarity over entity--citation edges, where an edge is treated as shared only when both the canonical entity and the resolved citation URL most closely following it match across two runs.
Edge-level agreement was near zero: across the 45 run pairs, the median edge Jaccard was 0.0073, with an interquartile range of 0.0019–0.0124 and a maximum of 0.0256. The median pair of runs shared only four entity--citation edges out of a union of 678.
We therefore do not plot edge overlap separately in Figure~\ref{fig:pairwise-strip}, as it would be visually indistinguishable from the citation row.
Substantively, the result indicates that repeated runs do not stably reproduce the same URL-level grounding structure.

Per-class Jaccard ranges over more than an order of magnitude, with Geopolitical actors and threat actors showing the highest median agreement, consistent with their being drawn from small, well-known closed vocabularies.
Security products, software targets, and cybersecurity techniques cluster near zero, which is consistent with their being drawn from large open vocabularies with long tails.
The overall entity row sits between these extremes and is dominated, in aggregate, by the long-tailed classes that contribute the most distinct items per run.

\begin{figure}[t]
	\centering
	\includegraphics[width=0.90\linewidth]{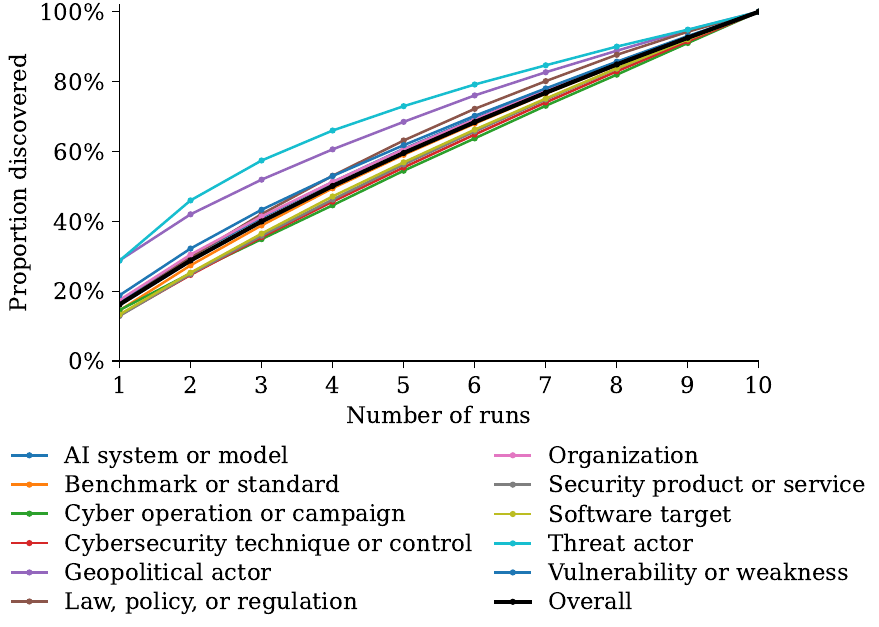}
	\caption{
    \textbf{Discovery curves for canonical entity coverage across repeated runs.} For each entity class, we randomly order the ten reports and compute the cumulative proportion of distinct canonical entities observed after $k$ runs. Curves that flatten quickly would indicate convergence towards a stable observed entity set; the mostly near-linear curves indicate that additional runs continue to surface previously unseen entities.}
	\label{fig:discovery}
\end{figure}

As an alternative way to visualize the diverging pattern of entity coverage, Figure~\ref{fig:discovery} shows the discovery curves for the entity classes.
Most curves appear approximately linear over the range we observe, with no visible inflection by the tenth run, with the exception of the categories Threat actor and Geopolitical actor.

\section{Concluding remarks}

Our experiment shows that a single commercial deep research agent, queried ten times with an identical prompt, produces reports with low but structured overlap.
The median overall entity Jaccard between runs is approximately $0.12$, the median citation Jaccard is just over $0.01$ and the median entity--citation edge overlap lower still, with a median Jaccard of $0.0073$.
The discovery curves do not visibly saturate within ten runs, indicating that additional runs continue to surface previously unseen entities.

The central finding is an asymmetry between referents and sources.
Repeated runs are more likely to mention the same organizations, geopolitical actors, threat actors, or AI models than they are to cite the same URLs in support of those mentions.
This suggests that the system has some stability in what it treats as topically relevant, but much less stability in the source trail through which those topics are retrieved and presented.

The research question related to entity–citation edges resolves negatively.
Exact edge agreement requires two runs to reproduce both the same canonical entity and the same resolved citation URL in support of it.
Because citation overlap is already extremely low, the resulting edge overlap is also near zero.

Finally, consistency is strongly class-dependent.
Geopolitical actors and threat actors show the highest pairwise agreement, while security products, software targets, and cybersecurity techniques cluster near zero.
This is consistent with the hypothesis that classes drawn from relatively bounded vocabularies are more stable than classes drawn from large, open-ended technical vocabularies.
Self-consistency is therefore not a single number that characterizes a DRA in isolation, but the result of an interaction among the system, the prompt, the domain, and the entity class being measured.

\section*{Limitations}

Several limitations impact the claims of this work.

\paragraph{Single system, prompt and domain.} We study one commercial DRA on one prompt in one domain. We make no claim that the magnitudes reported here generalize to other DRAs, other prompts, or other domains. Cybersecurity was chosen due to a hypothesis that it would produce a mix of relatively stable named entities and rapidly evolving long-tail entities, but other domains might show different patterns.

\paragraph{Conflation of retrieval-side and generation-side stochasticity.} Our experimental design does not separate variance arising from different web pages being retrieved across runs from variance arising from the same retrieved evidence being summarized differently due to limitations of the commercial API.

\paragraph{Web drift and URL stability.} All ten runs were conducted within a short time window, but we cannot rule out that some inter-run variance reflects genuine changes in the indexed web rather than stochasticity of the agent. Because entity–citation edge equality depends on exact resolved URL matches, the edge results should be interpreted as URL-level grounding stability rather than semantic evidence stability. Two different URLs may support similar claims, and the same underlying source may sometimes be reachable through different URL forms.

\paragraph{Extraction and canonicalization noise.} Our pipeline relies on LLM-based entity extraction and deduplication stages. We did not conduct a full annotation-quality audit of either stage. Spot checks suggest minor inconsistencies in granularity, for example whether versioned variants of the same model are merged into a single canonical entity or retained separately. This was likely a result of underspecified prompts to the deduplication software library. Such errors likely affect the exact numeric Jaccard values, especially for entity classes with many near-duplicate technical names but we do not have evidence that they would explain the overall patterns.

\bibliography{references}

\begin{thebibliography}{8}
\providecommand{\natexlab}[1]{#1}
\providecommand{\url}[1]{\texttt{#1}}
\expandafter\ifx\csname urlstyle\endcsname\relax
  \providecommand{\doi}[1]{doi: #1}\else
  \providecommand{\doi}{doi: \begingroup \urlstyle{rm}\Url}\fi

\bibitem[Asirvatham et~al.(2026)Asirvatham, Mokski, and
  Shleifer]{asirvathamGPTMeasurementTool2026}
Hemanth Asirvatham, Elliott Mokski, and Andrei Shleifer.
\newblock {GPT} as a {Measurement} {Tool}, February 2026.
\newblock URL \url{https://www.nber.org/papers/w34834}.

\bibitem[{Google}(2025)]{langextract111}
{Google}.
\newblock {LangExtract}, November 2025.
\newblock URL \url{https://github.com/google/langextract}.

\bibitem[Gupta et~al.(2026)Gupta, Chatterjee, Haas, Tao, Wang, Liu, Oiwa,
  Gribovskaya, Ackermann, Blitzer, Goldshtein, and
  Das]{guptaDeepSearchQABridgingComprehensiveness2026}
Nikita Gupta, Riju Chatterjee, Lukas Haas, Connie Tao, Andrew Wang, Chang Liu,
  Hidekazu Oiwa, Elena Gribovskaya, Jan Ackermann, John Blitzer, Sasha
  Goldshtein, and Dipanjan Das.
\newblock {DeepSearchQA}: {Bridging} the {Comprehensiveness} {Gap} for {Deep}
  {Research} {Agents}, January 2026.
\newblock URL \url{http://arxiv.org/abs/2601.20975}.
\newblock arXiv:2601.20975 [cs.CL].

\bibitem[Huang et~al.(2025)Huang, Chen, Zhang, Li, Zhou, Fang, Yang, Li, Shang,
  Xu, Hao, Shao, and Wang]{huangDeepResearchAgents2025}
Yuxuan Huang, Yihang Chen, Haozheng Zhang, Kang Li, Huichi Zhou, Meng Fang,
  Linyi Yang, Xiaoguang Li, Lifeng Shang, Songcen Xu, Jianye Hao, Kun Shao, and
  Jun Wang.
\newblock Deep {Research} {Agents}: {A} {Systematic} {Examination} {And}
  {Roadmap}, September 2025.
\newblock URL \url{http://arxiv.org/abs/2506.18096}.
\newblock arXiv:2506.18096 [cs.AI].

\bibitem[Min et~al.(2023)Min, Krishna, Lyu, Lewis, Yih, Koh, Iyyer,
  Zettlemoyer, and Hajishirzi]{minFActScoreFinegrainedAtomic2023}
Sewon Min, Kalpesh Krishna, Xinxi Lyu, Mike Lewis, Wen-tau Yih, Pang~Wei Koh,
  Mohit Iyyer, Luke Zettlemoyer, and Hannaneh Hajishirzi.
\newblock {FActScore}: {Fine}-grained {Atomic} {Evaluation} of {Factual}
  {Precision} in {Long} {Form} {Text} {Generation}, October 2023.
\newblock URL \url{http://arxiv.org/abs/2305.14251}.
\newblock arXiv:2305.14251 [cs].

\bibitem[Singh et~al.(2025)Singh, Ehtesham, Kumar, and
  Khoei]{singhAgenticRetrievalAugmentedGeneration2025}
Aditi Singh, Abul Ehtesham, Saket Kumar, and Tala~Talaei Khoei.
\newblock Agentic {Retrieval}-{Augmented} {Generation}: {A} {Survey} on
  {Agentic} {RAG}, February 2025.
\newblock URL \url{http://arxiv.org/abs/2501.09136}.
\newblock arXiv:2501.09136 [cs].

\bibitem[Zhai et~al.(2026)Zhai, Stengel-Eskin, Patil, and
  Leqi]{zhaiEvaluatingStochasticityDeep2026}
Haotian Zhai, Elias Stengel-Eskin, Pratik Patil, and Liu Leqi.
\newblock Evaluating {Stochasticity} in {Deep} {Research} {Agents}, February
  2026.
\newblock URL \url{http://arxiv.org/abs/2602.23271}.
\newblock arXiv:2602.23271 [cs].

\bibitem[Zhang et~al.(2025)Zhang, Li, Zhang, Jia, Wang, Guo, Liu, and
  Zhao]{zhangDeepResearchSurvey2025}
Wenlin Zhang, Xiaopeng Li, Yingyi Zhang, Pengyue Jia, Yichao Wang, Huifeng Guo,
  Yong Liu, and Xiangyu Zhao.
\newblock Deep {Research}: {A} {Survey} of {Autonomous} {Research} {Agents},
  August 2025.
\newblock URL \url{http://arxiv.org/abs/2508.12752}.
\newblock arXiv:2508.12752 [cs].

\end{thebibliography}

\end{document}